# GPU based GMM segmentation of Kinect data

Abdenour amamra, Tarek mouats and Nabil aouf

Cranfield University, Department of informatics and systems engineering, Shrivenham, wilts, SN6 8LA, UK

{a.amamra, t.mouats, n.aouf}@cranfield.ac.uk

*Abstract* - This paper presents a novel approach for background/foreground segmentation of RGBD data with the Gaussian Mixture Models (GMM). We first start by the background subtraction from the colour and depth images separately. The foregrounds resulting from both streams are then fused for a more accurate detection. Our segmentation solution is implemented on the GPU. Thus, it works at the full frame rate of the sensor (30fps). Test results show its robustness against illumination change, shadows and reflections.

## I. INTRODUCTION

Nowadays, real-time object tracking has become one of the largest studied subjects in robotics and computer vision domains. To track moving objects, we first need to detect them within the image. This detection requires a tool able to extract foreground pixels based on a background model. Hence arises the need to decide whether a new frame contains foreground regions or not. In addition, where the detected objects are located if there are any.

Several problems should be tackled by a good background removal algorithm. Such algorithm should be robust against non-stationary backgrounds like waving trees, sudden illumination changes and camouflage. Most of the current algorithms in the literature are able to decently cope with all the issues listed above. However, robustness and processing time constraints are critical requirements for a real-time tracking solution. To this end, we propose a new approach based on a well-established foreground/background segmentation algorithm (GMM) that uses two modalities (colour and depth images) to extract foreground regions. Furthermore, we benefit from the growing power of the GPUs to handle the relatively large amount of data in real-time.

## II. RELATED WORK

Many works from the literature addressed the possibility to jointly use the depth and colour images for foreground/background segmentation. Cristani et al. [1] proposed an overview on background subtraction solutions for mono as well as stereo cameras. Abramoff et al. [2] used a stereo pair of cameras for an automatic segmentation algorithm. Gordon et al. [3] added the disparity information to the GMM for background modelling. In their approach, the authors found that the combination of stereo and colour data helps to overcome the classic problems of colour segmentation. However, stereo images themselves result from a pair of RGB images. Thus, they hold weaknesses toward illumination change and shadow. Lee et al. [4] proposed a GPU implementation of Stauffer and Grimson's algorithm. Friedman and Russell first proposed GMM approach for background removal [5]. Few years later, innovations were introduced to the basic model by Stauffer and Grimson [6]. Their paper is often considered as the reference of the GMM for background/foreground segmentation. Lee et al. [4] later proposed a GPU implementation of Stauffer and Grimson's algorithm.

To the best of the author's knowledge, this work is the first time when GMM background subtraction algorithm is implemented on the GPU for joint colour and depth background removal. Our approach is innovative in two points: the data is streamed by a cheap depth camera. This data is both the colour image and the range map. Hence, two different modalities independent from each other. It also leverages the GPU to reduce the computational load.

## III. GAUSSIAN MIXTURE MODELS FOR RGBD DATA

### A. Background modelling

The GMM is a parametric probability density function represented as a weighted sum of Gaussian densities [7]. Foreground detection generally follows the steps listed below:

- Modelling the values of a particular pixel as M ($3 \leq M \leq 5$) mixture of Gaussians G{$\mu_i$, $\sigma_i$, $w_i$}; $1 \leq i \leq M$ Fig 1(a). Where each distribution is characterised by a mean $\mu_i$ and a variance $\sigma_i^2$ and weighting factor $w_i$ to define the importance of each Gaussian. The $w_i$ are positive and add up to 1.
- Determining the Gaussian that corresponds to the background model based on the mean and the variance of each of the **M** distributions.
- The foreground is then defined by the pixels that do not fit the background.
- Updating the Gaussians with the newly detected foreground pixels.
- The pixel values that do not match one of the **M** background Gaussians are grouped to form a foreground blob.

For every new frame $f$, the GMM algorithm computes the distance between the pixel $f(u,v)$ and each of the means $\mu_i(u,v)$ characterising the **M** Gaussians Fig 1(b). We test the new pixel against the highly weighted distribution and vice versa. If the new pixel does not match any of the recorded distributions then it is a foreground element. The background model is then updated.

### B. GMM on RGBD data

Light intensity images are naturally sensitive to the illumination change and the shadow.

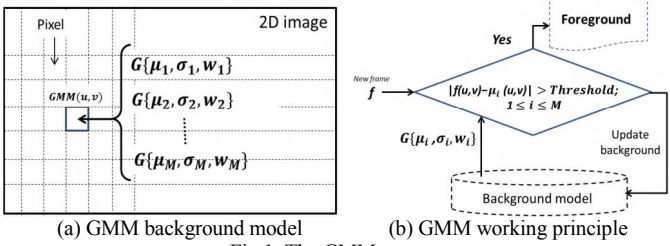

(a) GMM background model    (b) GMM working principle
Fig 1. The GMM structure

On the other hand, the depth data is proven to be robust against the lighting of the scene. The concept of fusing the depth and the RGB images for background removal can be approached in two different ways: The first is the **augmentation** of the three colour channels with a forth component for depth. The experimental test results showed that the resulting image still suffers from the classic artefacts of RGB based GMM. The contribution of the depth data in this segmentation model is weakened by the three other intensity bytes. Hence, the resulting image is almost the same as the one without considering depth information Fig 2.

On the other hand, in our work we have completely **separated** the two modalities during the background removal phase. In other word, we apply the GMM on the RGB image and the depth map independently. The resulting binary foreground images are combined to produce the final result Fig 3.

*C. RGBD Background fusion*

Before mixing the results of the two independent segmentations, we transform the depth foreground image to the colour space. This mapping is required because the two modalities do not share the same reference frame. Afterwards, we adapt the following the algorithm in List 1 to fuse the two binary foreground images:

```
2D binary image f'i_RGB, f'i_D , I'Forg;//foreground
2D integer array cpt;
For every pixel (u,v)
 If (f'i_RGB(u,v) == f'i_D(u,v))
   I'Forg(u,v) = f'i_D(u,v)
   cpt(u,v)=0
 Else //different
   If (cpt(u,v) == +3)
     I'Forg(u,v) = f'i_rgb(u,v)
     cpt(u,v)=0
   Else if (cpt(u,v) == -3)
     I'Forg(u,v) = f'i_d(u,v)
     cpt(u,v)=0
   Else If (I'Forg(u,v) == f'i_rgb(u,v))
     cpt(u,v) = cpt(u,v)+1;
   Else
     cpt(u,v) = cpt(u,v)-1;
 end
end
```
List 1. RGBD foreground fusion algorithm

When the two responses are different, then a decision should be taken regarding the fused outputs. To this end, we apply this rule: The pixel keeps the same state (as it was before the confusion occurs), as long as no three successive frames sharing the same state List 1. This assumption is motivated by the fact that we noticed that permanent incoherence between the colour and depth GMMs is considered as a noise factor disturbing the current state.

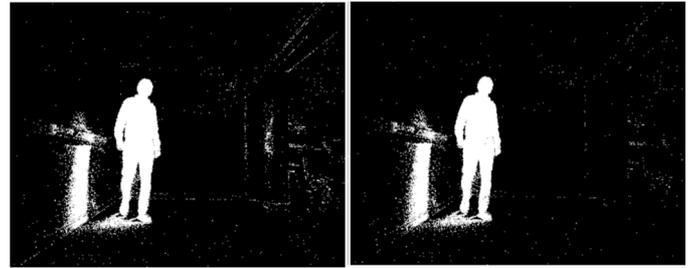

(a) RGB    (b) RGB+D
Fig 2. RGBD segmentation (the depth as a forth component)

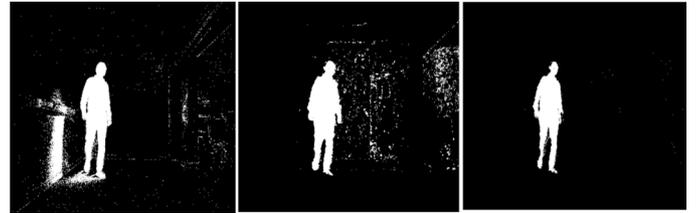

(a) RGB    (b) RGB+D    (c) RGB&D
Fig 3. RGBD segmentation (the depth as a separate component)

IV. GPU ACCELERATION OF THE GMM

Naturally, image data can be processed in parallel on the GPU by associating a thread to each pixel. In our case, we only use VGA resolution (640×480) colour and depth data. When we naively adapt such approach (thread to each pixel), the maximum achievable frame rate on the GPU was **18fps**. Consequently, other optimisation strategies should be taken into account to fully exploit all the available hardware capability. Basically, the design of heterogeneous algorithms aims a higher occupancy of the processors and a full usage of the bandwidth when exchanging data within the GPU and between the central memory (RAM) and the global memory of the GPU (GMGPU). To this end, we focus on: **running asynchronous transfers** of the following frame ($f_{i+1}$) from the RAM to the GMGPU, and the already available result from GMGPU to the RAM ($f_{i-1}$). Simultaneously, the current frame ($f_i$) is processed on the device Fig 4(a). We perform the exchange of data on the bus when the GPU processes the current frame.

*Memory coalescing* is another optimisation that significantly helps to increase the probability of threads in the same warp to feed from neighbouring memory emplacements. Appropriately organising the data in the device memory allows such contiguous access to automatically happen. Fig 4(b) illustrates what happens when a thread calls a given cell in the global memory. The GPU automatically loads the content of the adjacent cells the internal design of the device assumes that it is highly probable that the neighbouring data will be sooner requested as well [8]. The programmer should consider this fact by transforming the array of structures data to a structure of arrays. We therefore apply some changes on our RGB image. In other words, the RGB data is transformed to three arrays each corresponding to the one of the channels (Red, Green and Blue) Fig 5(a). On the other hand, the depth map does not need to be reorganised as it naturally holds one component, which is the actual depth reading. We apply the same change on all the parameters of the GMM by splitting them on contiguous arrays Fig 5(b), (c).

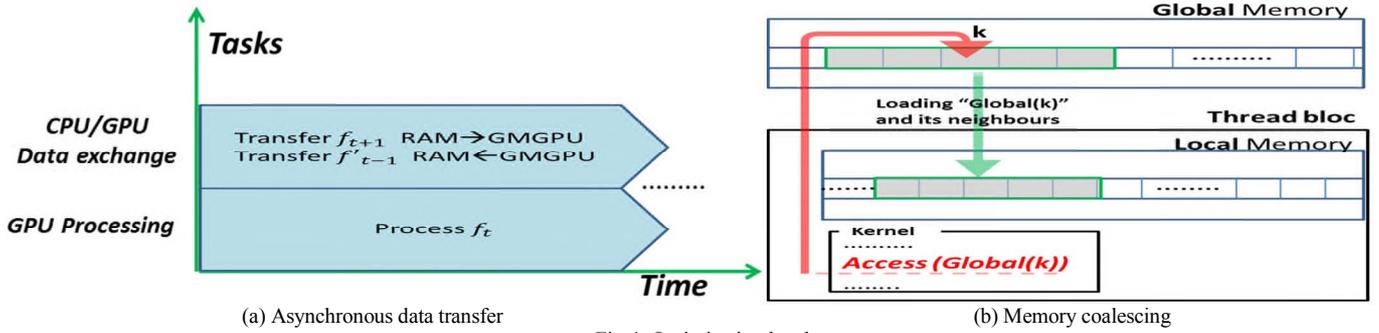

(a) Asynchronous data transfer  (b) Memory coalescing
Fig 4. Optimisation levels

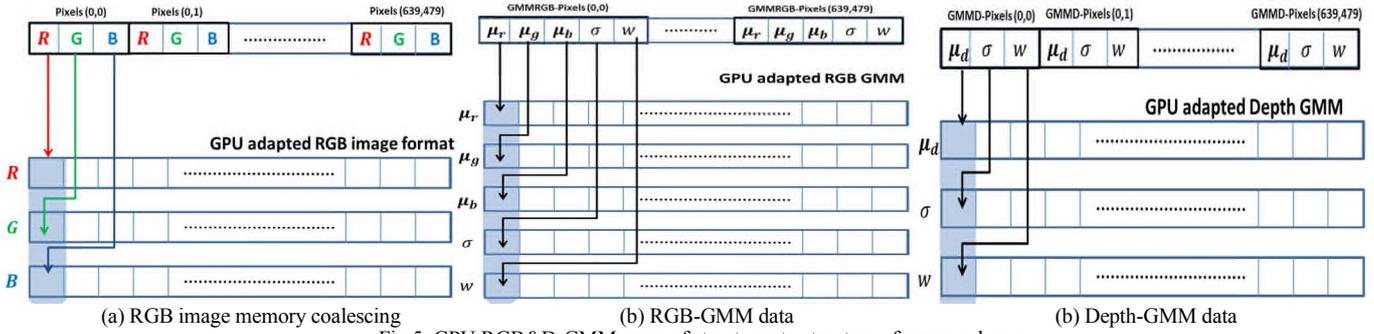

(a) RGB image memory coalescing  (b) RGB-GMM data  (b) Depth-GMM data
Fig 5. GPU RGB&D-GMM array of structures to structure of arrays scheme

## V. EXPERIMENTAL RESULTS

### A. RGBD-GMM

To validate our finding, we conducted two experiments on an ordinary scene (with two major illumination changes and a gently moving background) and a challenging scene where the illumination keeps changing during the whole capture Fig 6. The evaluation parameter $F_1$ [9] is computed based on the false and the true positives and negatives $(TP, FP, TN, FN)$. the *recall* is the true positive rate:

$$R = TP/(TP + FN) \qquad (1)$$

And the ***precision*** which is the ratio between the number of correctly detected pixels and the total number of pixels marked as foreground.

$$P = TP/(TP + FP) \qquad (2)$$

Accuracy metric $\boldsymbol{F_1}$ combines the *precision* and *recall* to effectively evaluate the accuracy of segmentation

$$F_1 = 2\frac{PR}{P+R} \qquad (3)$$

$F_1$ is a good tool to evaluate the robustness. The higher the value of this metric becomes, the better the performance will be.

The graphs in Fig 7(a), (b) illustrate the behaviour of $F_1$ in two different experiments where we plot $F_1$ for all three responses (**RGB**, **Depth**, and the fusion of the two **RGB&D**). In Fig 7(a) $F_1$ characterising the colour image undergoes two major drops which correspond to important perturbations mainly due to sudden illumination change (Red). Whereas, the depth map suffers less from problems of illuminations (Blue) its $F_1$ mostly remains above 0.90. $F_1$ for the fused outputs (Green) is above 0.97 which clearly shows the robustness of our segmentation algorithm. On the other hand, in Fig 7(b) all three methods suffer from perturbations along the whole period of capture. RGB image remains the most affected though. The depth map is also affected because the background was very dynamic but the latter kept the effectiveness. $F_1$ mostly remains above 0.80 for the fused outputs although the highly challenging situation.

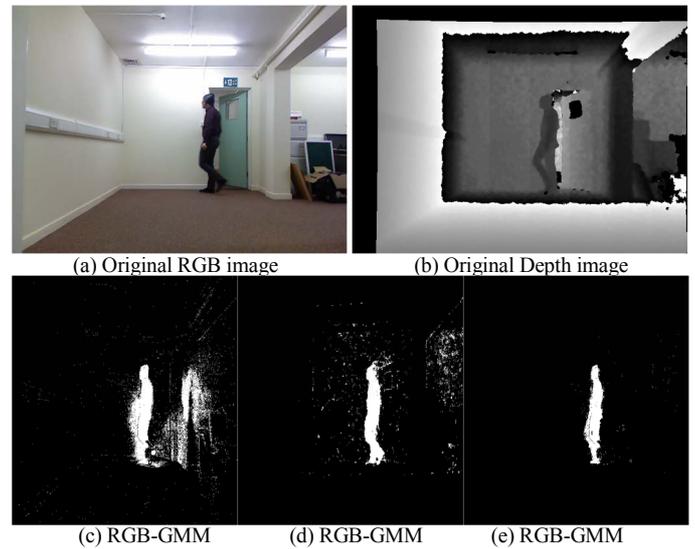

(a) Original RGB image  (b) Original Depth image

(c) RGB-GMM  (d) RGB-GMM  (e) RGB-GMM
Fig 6. RGB&D-GMM

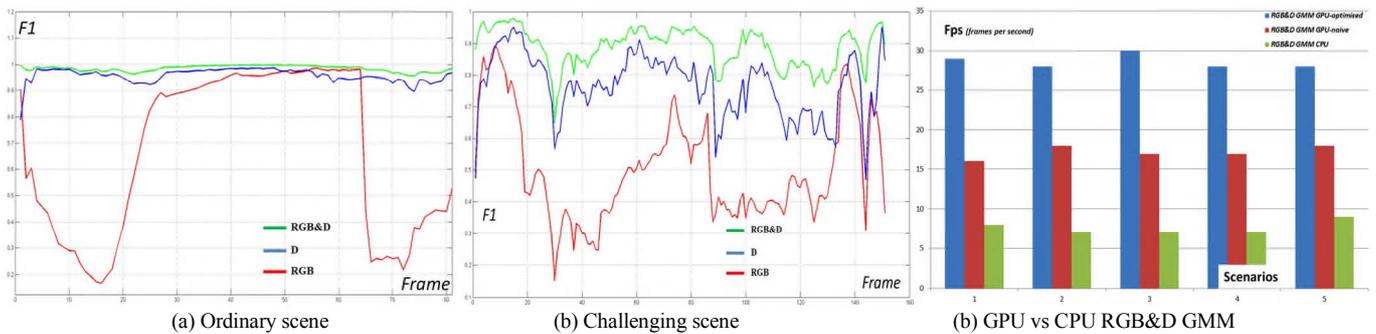

(a) Ordinary scene  (b) Challenging scene  (b) GPU vs CPU RGB&D GMM
Fig 7. RGB&D-GMM results

## B. Processing time metrics

The implementation of our algorithm in the GPU allowed us to fully benefit from the available frame rate of Kinect sensor (30fps). The latter remains just below 30fps for all the five experimental scenarios. The effectiveness of the GPU can be clearly seen (blue and red bars in Fig 7(c), against the green bars). More importantly, the improvements after considering the memory coalescing and the data exchange optimisations increased the final frame rate of the segmentation algorithm to achieve almost 30fps.

## VI. CONCLUSION AND FUTURE WORKS

We presented an innovative approach to couple RGB and Depth images for real-time background removal. Based on native GMM (implemented on the CPU), we designed a parallel algorithm which benefits from asynchronous data exchange between the host (CPU) and the device (GPU). In addition, we organised our data structures in the GPU to permit a higher memory coalescing. Finally, we validated our findings on two real scenarios in different conditions. The results were very promising, although we only applied background subtraction. This work opens a new perspective on using RGBD data to tackle the classic problems of RGB imagery. Our next step is to jointly using multiple RGBD sensors to detect moving robots in an indoor scene based on the background subtraction we already achieved.